\documentclass[letterpaper, 10 pt, journal, twoside]{IEEEtran}

\usepackage{graphicx}
\usepackage{epsfig} 
\usepackage{mathptmx}
\usepackage{times} 
\usepackage{amsmath}
\usepackage{amssymb}  
\usepackage{tabularx}
\usepackage{xcolor}
\usepackage{here}
\usepackage{multirow}

\renewcommand\fbox{\fcolorbox{white}{white}}

\begin{document}
\title{CodeMapping: Real-Time Dense Mapping for Sparse SLAM using Compact Scene Representations}

\author{Hidenobu Matsuki, Raluca Scona, Jan Czarnowski and Andrew J. Davison
\thanks{Manuscript received: February, 24, 2021; Revised May, 29, 2021; Accepted June, 28, 2021.}
\thanks{This paper was recommended for publication by Editor Javier Civera upon evaluation of the Associate Editor and Reviewers' comments.}
\thanks{Research presented in this paper has been supported by Dyson Technology Ltd. }
\thanks{The authors are with Dyson Robotics Laboratory, Imperial College London, United Kingdom {\tt\small h.matsuki20@imperial.ac.uk}}%
\thanks{Digital Object Identifier (DOI): see top of this page.}
}

\markboth{IEEE Robotics and Automation Letters. Preprint Version. June, 2021}
{Matsuki \MakeLowercase{\textit{et al.}}: CodeMapping} 

\maketitle

\begin{abstract}
We propose a novel dense mapping framework for sparse visual SLAM systems which leverages a compact scene representation.	
State-of-the-art sparse visual SLAM systems provide accurate and reliable estimates of the camera trajectory and locations of landmarks. While these sparse maps are useful for localization, they cannot be used for other tasks such as obstacle avoidance or scene understanding.
In this paper we propose a dense mapping framework to complement sparse visual SLAM systems which takes as input the camera poses, keyframes and sparse points produced by the SLAM system and predicts a dense depth image for every keyframe. We build on CodeSLAM~\cite{bloesch2018codeslam} and use a variational autoencoder (VAE) which is conditioned on intensity, sparse depth and reprojection error images from sparse SLAM to predict an uncertainty-aware dense depth map. The use of a VAE then enables us to refine the dense depth images through multi-view optimization which improves the consistency of overlapping frames.
Our mapper runs in a separate thread in parallel to the SLAM system in a loosely coupled manner. This flexible design allows for integration with arbitrary metric sparse SLAM systems without delaying the main SLAM process. Our dense mapper can be used not only for local mapping but also globally consistent dense 3D reconstruction through TSDF fusion. We demonstrate our system running with ORB-SLAM3 and show accurate dense depth estimation which could enable applications such as robotics and augmented reality.
\end{abstract}


\section{INTRODUCTION}
\label{sec:introduction}
\begin{figure}[tb]
    \centering
		\includegraphics[width=8.0cm]{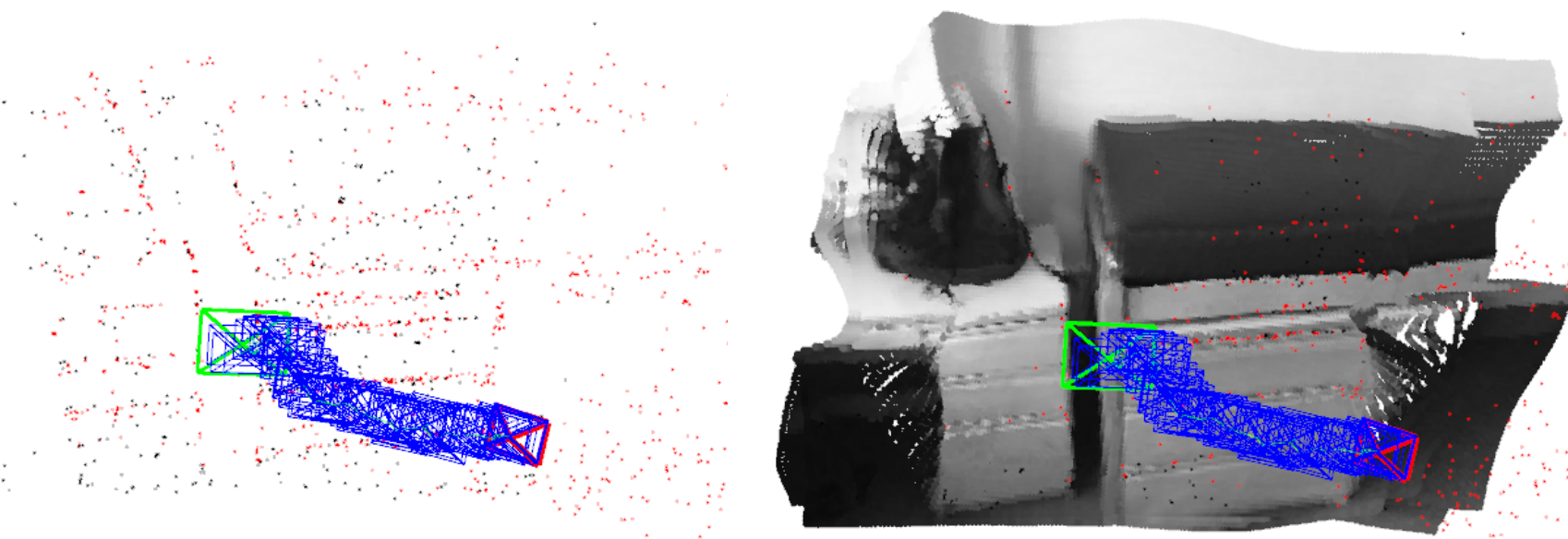}

		\label{top}
	
        \centering
		\includegraphics[width=6.8cm]{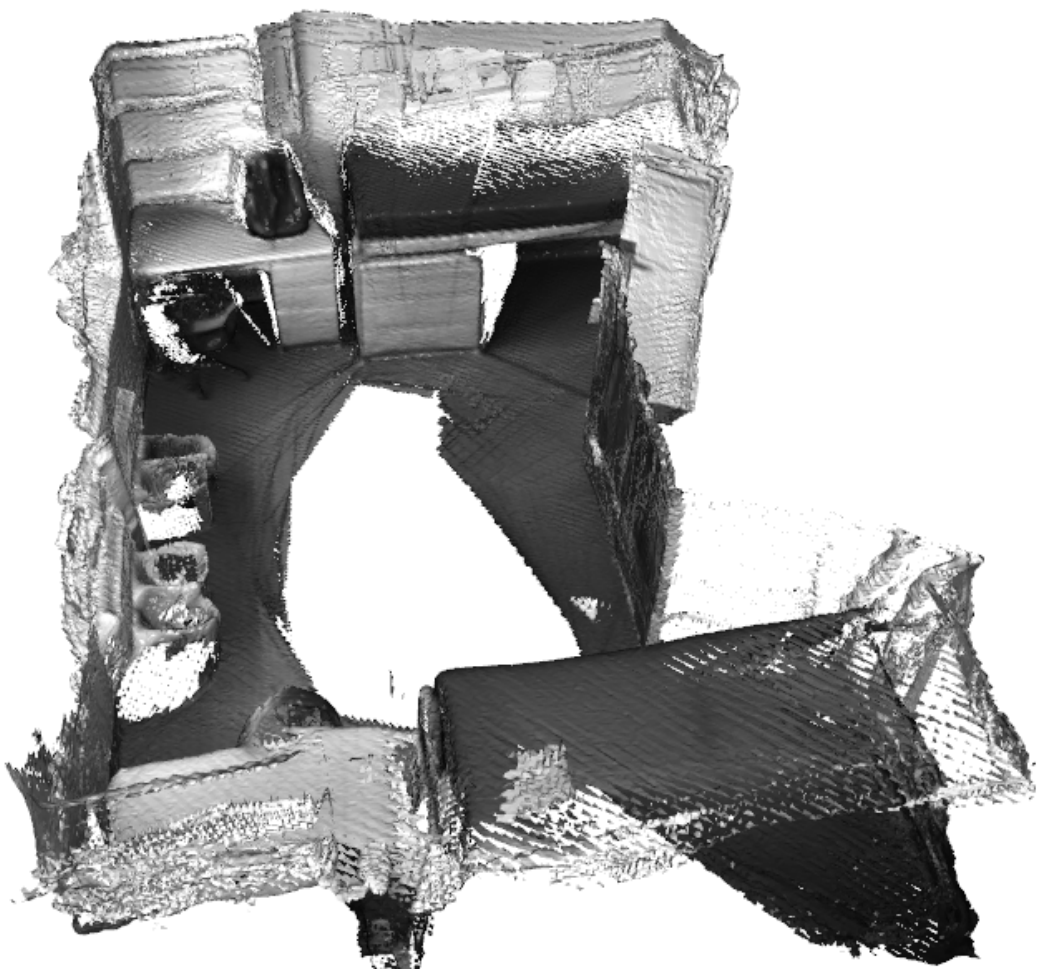}
				\caption{{\bf{Top: }} Real-time dense mapping result (\textit{left}: sparse SLAM, \textit{right}: ours). Our dense mapper completes the sparse scene geometry and refines it by sliding-window multi-view optimization {\bf{Bottom: }} An example of global 3D reconstruction by TSDF fusion of our mapping system. Our method can be applied to arbitrary metric sparse SLAM systems.
				}
        
\end{figure}
\IEEEPARstart{V}ision based Simultaneous Localization and Mapping (SLAM) has made remarkable progress over the past 20 years and has a wide range of commercial applications such as robot navigation and AR/VR. Prior research has shown that the key elements for SLAM accuracy are informative point selection (typically corners or edges) and joint optimization of the camera poses and landmark locations (bundle adjustment). Thanks to these, sparse SLAM produces more accurate results compared to dense methods which decouple map building and camera pose estimation \cite{platinsky2017monocular}\cite{DSO}\cite{campos2020orb}. When used in practical applications, sparse SLAM is often fused with other sensors to improve robustness and also provide metric scale \cite{campos2020orb} \cite{houseago2019ko} . However, the sparse feature map is not visually informative and cannot directly be used for tasks such as collision-free motion planning or surface-aware AR. Several methods have attempted to perform dense mapping and camera localization simultaneously \cite{newcombe2011dtam}\cite{engel2014lsd}, but they were not as successful as the sparse methods because (1) dense image alignment is vulnerable to photometric noise and (2) the large number of scene parameters restricts real-time joint optimization. Therefore, instead of running dense SLAM, it can be more practical to use a separate algorithm to infer dense depth. 
This densification has commonly been done by Multi-View Stereo algorithms but in recent years many depth completion methods using deep learning have also been investigated and show promising performance \cite{ma2018sparse}\cite{zhang2018deep}\cite{xiong2020sparse}. 

In this paper we propose {\it{CodeMapping}}, a novel real-time dense mapping method which leverages the geometric prior information provided by sparse SLAM. We extend the compact scene representation proposed in CodeSLAM~\cite{bloesch2018codeslam} to complete sparse point sets. We train a VAE to predict a dense depth map and condition it on the corresponding gray-scale image, a sparse depth image from the observable features in the map and a reprojection error image, to enable uncertainty-aware depth completion. Consecutive depth maps are refined through multi-view optimization using fixed camera poses provided by the SLAM system, to improve their consistency and accuracy. Our system works in parallel to the SLAM system, subscribing to the estimated camera poses, sparse depth and reprojection error maps. This formulation enables flexible keyframe selection for multi-view dense bundle adjustment and real-time execution without interrupting the main SLAM process.

\section{RELATED WORK}
\label{sec:relatedwork}
\begin{figure}[H]
        \centering
		\includegraphics[width=8.5cm]{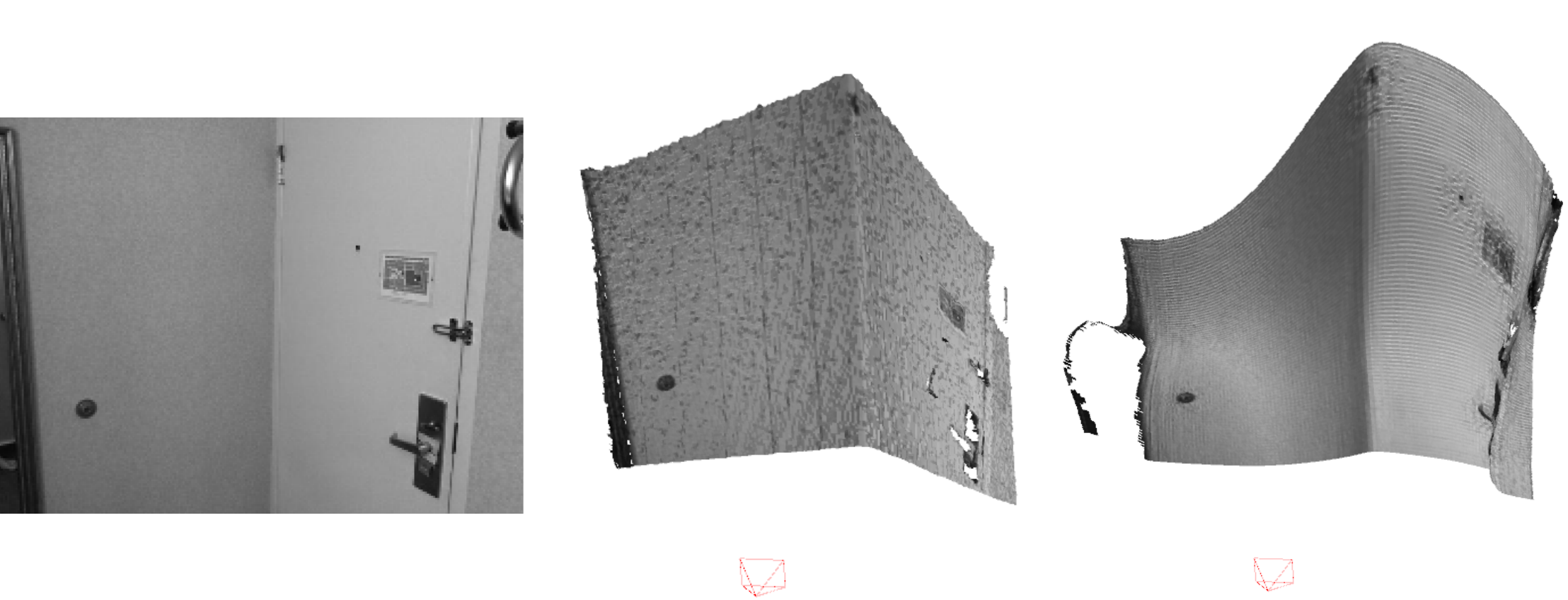}
				\caption{
    				An example of depth image prediction using DeepFactors \cite{czarnowski2020deepfactors} ({\bf{Left: }}Input image, {\bf{Middle:}} 3D visualization of depth from an RGBD sensor, {\bf{Right:}} 3D visualization of the depth image estimated with DeepFactors). We can clearly see that DeepFactors' prediction is incorrectly bent around the wall and the door. 
				}
		\label{fig:df}
        
\end{figure}

{\bf Visual SLAM:}
SLAM methods proposed in the early 2000s used a Kalman Filter to estimate the camera pose and the map\cite{monoslam}. However, since filtering-based SLAM includes all landmark positions and camera poses in a state space and updates all the variables at every time step, the method can not be applied to large scale scenes due to computational limitations. A major breakthrough was made in 2007 by PTAM~\cite{klein2007parallel}, a keyframe-based SLAM method which runs tracking and mapping in separate threads. Similarly, many current SLAM algorithms also use keyframes and apply multi-threading to separate real-time locally consistent tracking and slower globally consistent mapping into different threads. Various types of scene representations have been proposed for keyframe-based SLAM, but most current methods are using sparse 3D points. This is because joint optimization of poses and geometry is necessary for accurate performance and a sparse set of points ensures real-time operation. Dense geometry is desirable for scene understanding but due to the large number of parameters normally used to represent it, joint optimization in real-time has not been possible.
CodeSLAM~\cite{bloesch2018codeslam} and DeepFactors~\cite{czarnowski2020deepfactors} proposed a dense compact scene representation using VAE, but they predict a dense depth map purely from a gray-scale image which can be inaccurate in practice (Fig.~\ref{fig:df}).

\begin{figure*}[h]
        \centering
		\includegraphics[width=18.0cm]{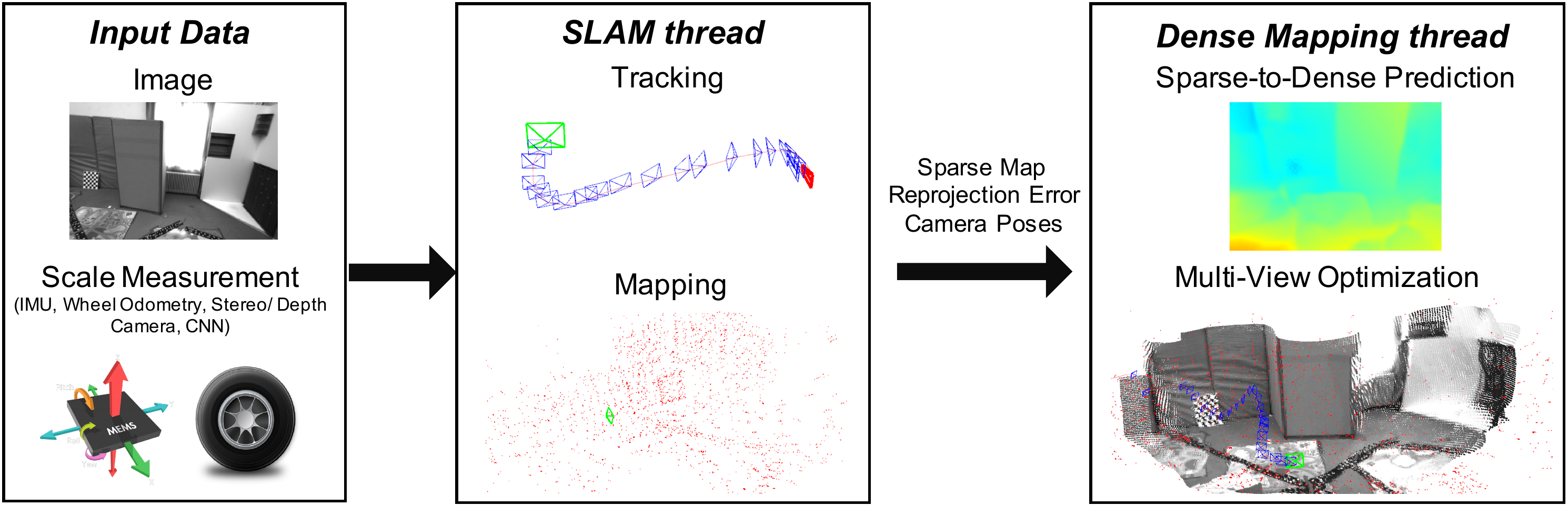}
				\caption{System overview of CodeMapping: the dense mapper subscribes to bundle-adjusted sparse point sets and camera poses to run dense prediction and multi-view optimization in a separate thread.}
		\label{fig:overview}
        
\end{figure*}

{\bf SLAM with Scale Estimation:}
In practice, Visual SLAM is rarely used without scale information. Metric scale is recovered using additional sensors or priors such as IMUs \cite{campos2020orb}\cite{mourikis2007multi}\cite{rosinol2020kimera}, wheel odometry \cite{houseago2019ko}, known camera height \cite{zhou2016reliable}, stereo/depth cameras \cite{campos2020orb}\cite{schops2019bad}. Recently, CNN-based depth predictions are used to infer metric scale for monocular visual SLAM \cite{cnn}\cite{yang2020d3vo}. 
Using scale information, camera trajectory estimates from sparse VO/SLAM systems exhibit errors smaller than a few percent of the total traveled distance. These systems can be used for a variety of tasks such as mobile robot navigation and position tracking for AR/VR. 

{\bf Depth Completion using Deep Learning:}
Reconstruction of dense depth from sparse points was classically formulated as an energy minimization problem \cite{ferstl2013image}. With the recent development of deep learning, learning-based depth completion methods have shown promising performance. For example, Ma~\emph{et al.}~\cite{ma2018sparse} proposed a color-guided encoder-decoder network and revealed that sparse-point guided depth prediction shows much better performance than purely image-based depth prediction. Zhang~\emph{et al.}~\cite{zhang2018deep} used predicted surface normals to densify Kinect depth images, and Xiong~\emph{et al.}~\cite{xiong2020sparse} applied a Graph Neural Network to depth completion. Cheng~\emph{et al.}~\cite{cheng2020s3cnet} proposed a semantic scene completion network for LiDAR point clouds for a large scale outdoor environment.
However, these networks are designed for single-view densification and do not actively address multi-view consistency when processing video streams. 
Recent techniques propose to estimate dense 3D reconstructions from posed images. For example, Atlas~\cite{murez2020atlas} directly predicts TSDFs and DeepMVS ~\cite{huang2018deepmvs} predicts disparity maps.  While both methods produce impressive reconstructions, they require significant amounts of GPU memory to run 3D CNNs in the case of Atlas and plane-sweeping volumes in DeepMVS. Regarding DeepMVS, the large structure of the network also makes this method more suited to offline applications.
Our approach is more modular and composed of a number of steps. The different modules can be used in different tasks depending on the run-time and memory requirements. For example, in time-critical tasks such as obstacle avoidance, only the single-view depth completion network may be sufficient. If a dense model is required, the multi-view optimization and TSDF fusion can be used at an increased computational cost.

{\bf Real-time Dense Mapping:}
The work most similar to ours is CodeVIO\cite{zuo2020codevio}, which uses a CodeSLAM-like VAE conditioned on an intensity image and a sparse depth map. Their system is tightly coupled with filtering based visual-inertial odometry and simultaneously estimates scene geometry and camera trajectory in the same state space. Compared to their system, our method has the following advantages: first, our network is conditioned not only on intensity and depth but also on the reprojection error of points, enabling the network to consider the confidence of SLAM points when estimating dense depth. Second, our system runs as a separate backend process and is loosely-coupled with a keyframe-based SLAM system. Keyframe-based SLAM can estimate a larger number of points in real-time when compared to filtering which is important for sparse-to-dense networks as more information provided to the network simplifies the learning task and can lead to more accurate results. Third, since keyframe-based sparse SLAM provides a globally consistent camera trajectory, our depth maps can be used not only for local mapping but also global depth map fusion.

\section{System Overview}

Fig.~\ref{fig:overview} shows an overview of our system. The system launches a SLAM process (tracking, local and global mapping threads) and a Dense Mapping process which continuously runs in parallel until the input data stream stops. In this work, we use ORB-SLAM3~\cite{campos2020orb} as a sparse SLAM system because it supports multimodal sensor input and shows state-of-the-art performance. We describe these components next.

\subsection{SLAM Thread}
{\bf Tracking:} Tracking uses sensor information to estimate the pose of the camera relative to the existing map. ORB-SLAM3 runs pose-only bundle adjustment which minimizes the reprojection error of matched features between the current image and the map. This process is done for every input frame in real-time and it also decides whether the current frame becomes a keyframe to be included in the map.


{\bf Mapping:} Mapping runs bundle adjustment for pose and geometry refinement. ORB-SLAM3 runs local and global bundle adjustment for this process. Local bundle adjustment uses only a small number of keyframes for optimization, while global bundle adjustment uses the whole graph. After every local bundle adjustment, the SLAM thread transfers data from a window of four keyframes to the Dense Mapping thread.
The four keyframes consist of the latest keyframe and the top three covisible keyframes based on the ORB-SLAM3's covisibility criteria. The keyframe data consists of respective camera poses, sparse depth and reprojection error images.
Each sparse depth map is obtained by projecting 3D landmarks on the keyframe and the reprojection error map is calculated by projecting each 3D landmark to all observed frames and computing the average distance to the corresponding matched ORB feature location. The reprojection error of newly registered keypoints with no matches from other frames are initialized with a large value (10).

\subsection{Dense Mapping Thread}
{\bf Sparse-to-Dense Prediction:} 
\label{Sparse-to-dense section}
When the dense mapping thread receives keyframe data from the SLAM thread, it checks whether the keyframe data has been previously processed. If not, the system runs the depth completion VAE using the the sparse depth and reprojection error input images to predict an initial dense depth map. The depth completion VAE also generates low dimensional latent codes for the predicted dense depth maps, which are used later in multi-view optimization. This is described in Sec. \ref{sec::CVAE} and the runtime of this process is provided in Table \ref{table:runtime}.
 
\begin{figure}[h]
    \centering
		\includegraphics[width=9.0cm]{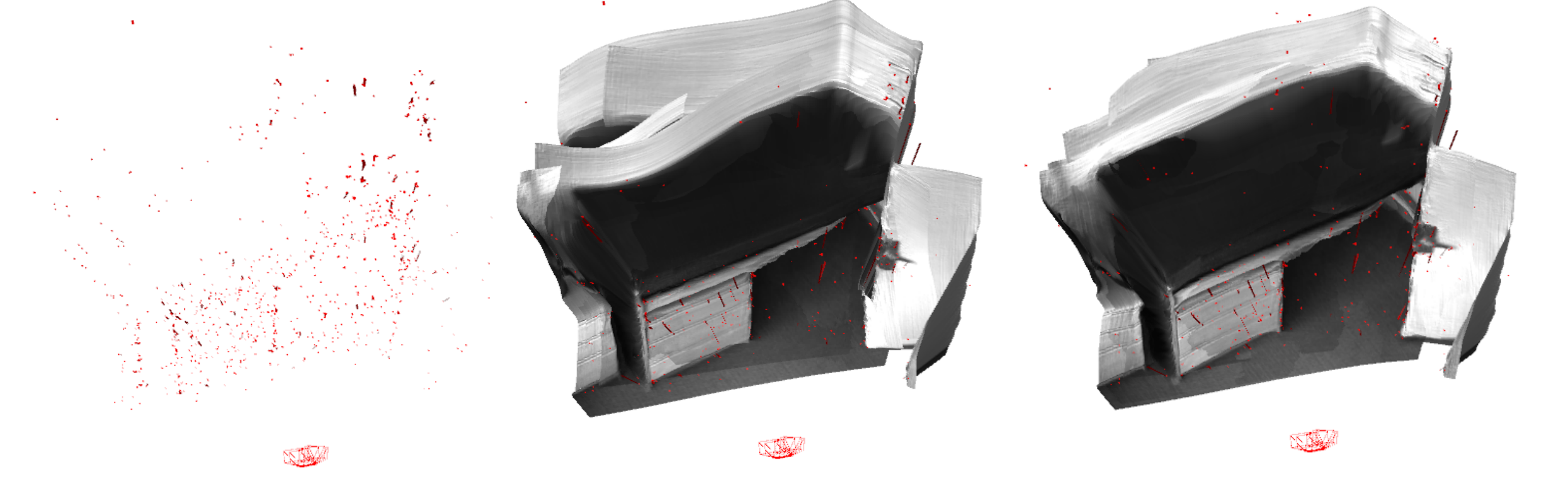}
		\caption{
    		An example of 5-view depth completion and multi-view optimization ({\bf{left:}} sparse SLAM points, {\bf{middle:}}  initial depth completion, {\bf{right:}} 5-view optimization). Multi-view optimization improves depth estimation accuracy especially in texture-less regions such as the white wall.
		}
		\label{fig:multiview}
\end{figure}

{\bf Multi-view Optimization:}
After an initial prediction, the depth maps are refined by sliding window multi-view optimization using the camera poses provided by the SLAM system (Fig.~\ref{fig:multiview}). We use the factor-graph based optimization proposed in DeepFactors, which considers different types of factors between overlapping frames.
We optimize only the per-frame codes, but not the camera poses as we assume the poses from sparse SLAM are already accurate enough and pixel-wise dense alignment is less likely to improve the accuracy further. We describe this in detail in Sec. \ref{section:multiviewOpt}.

\section{Sparse-to-Dense Prediction}
\label{sec::CVAE}
\subsection{Network Architecture}
\begin{figure}[h]
    \centering
    \fbox{
		\includegraphics[width=8.5cm, height=5.5cm]{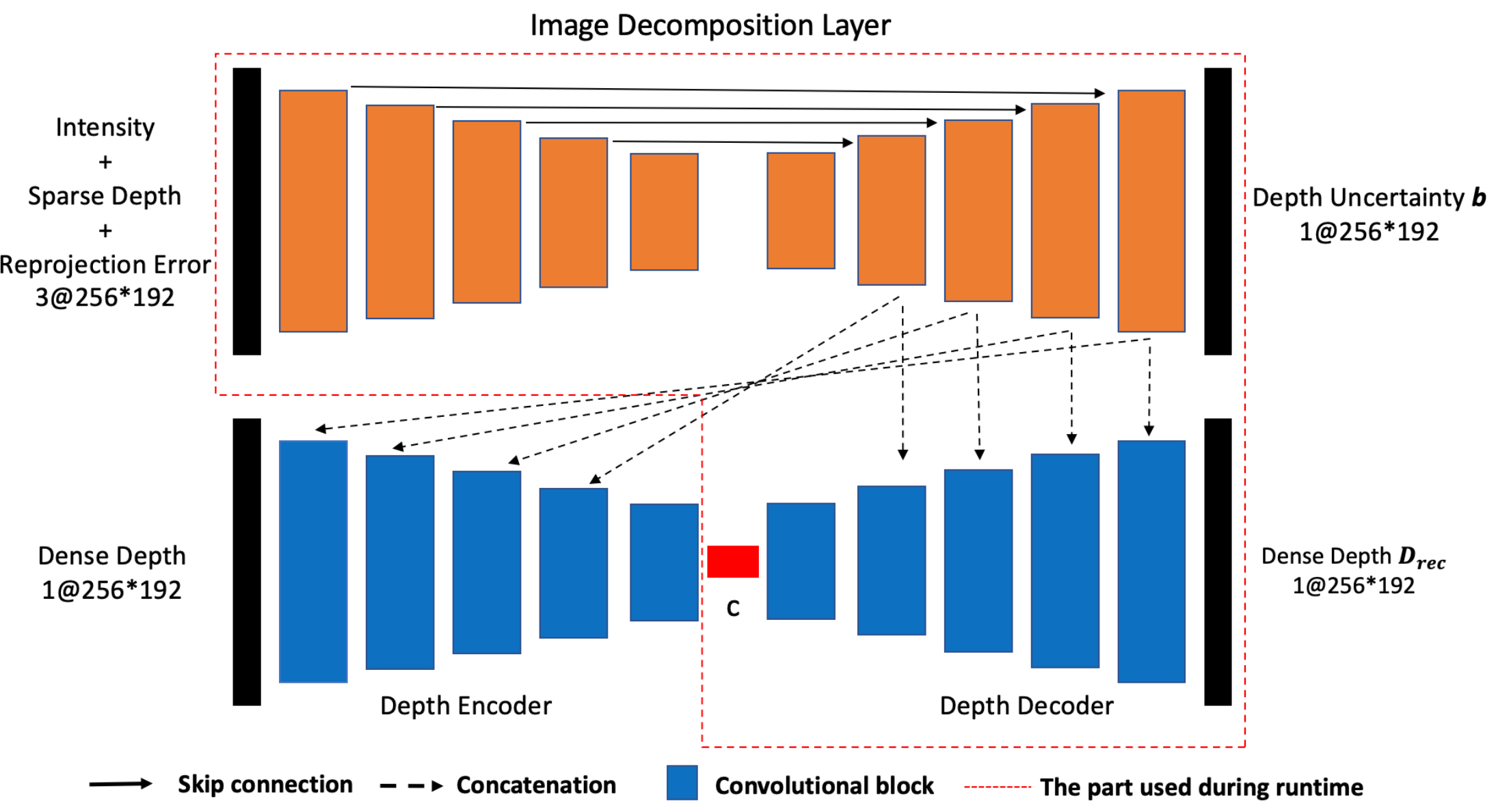}
	}
		\caption{
		Architecture of the proposed conditional VAE network. 
		}
		\label{fig:network}
	
\end{figure}

\begin{figure}[h]
    \centering
		\includegraphics[width=7.5cm]{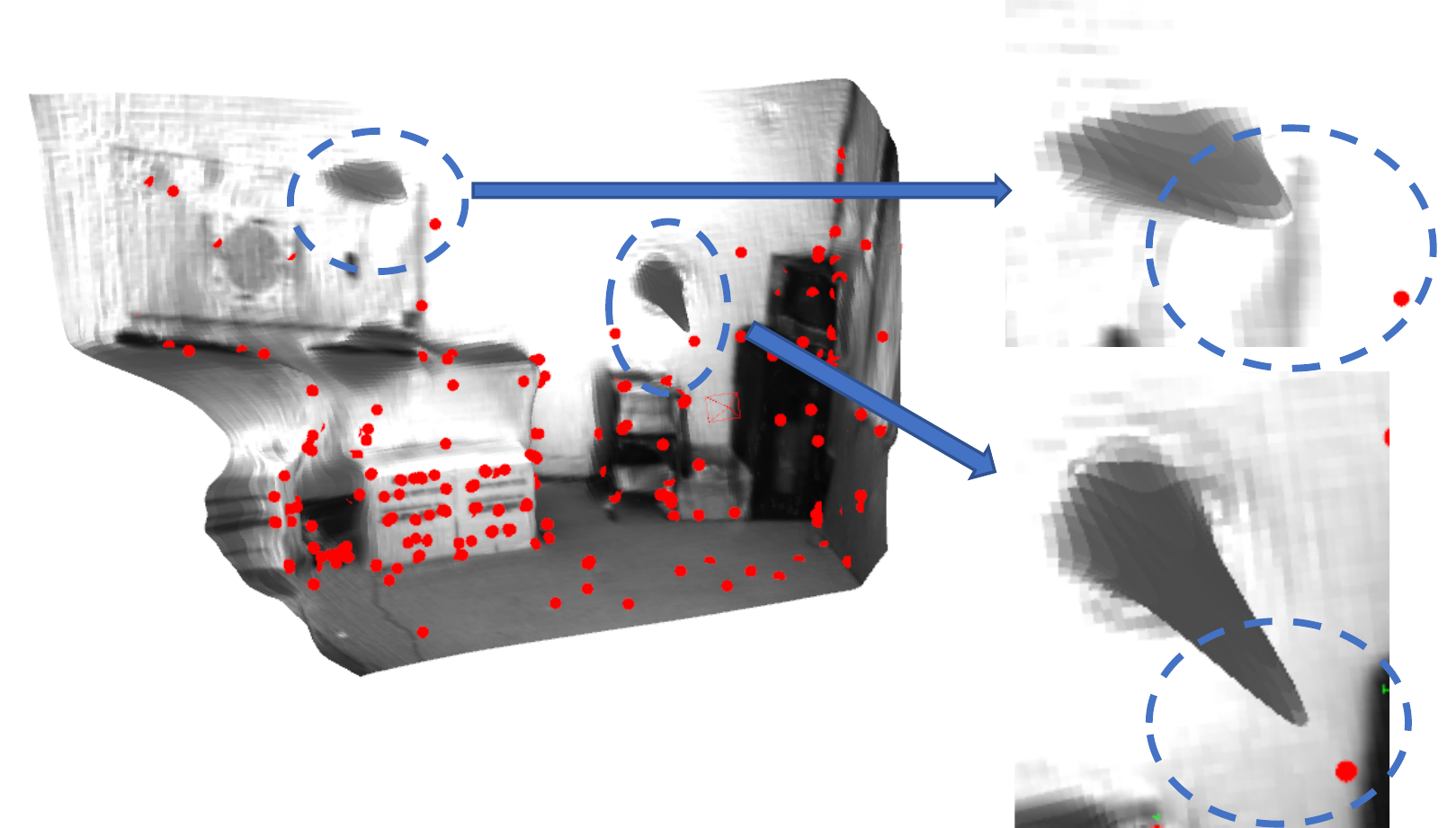}
		\caption{
		    Depth prediction without reprojection error conditioning. Red points are generated by sparse SLAM and the network completes the dense geometry. As highlighted in blue circles, the prediction is severely affected by outlier SLAM points.
		}
		\label{fig:rep_demo}
	
\end{figure}

We extend the conditional variational autoencoder from DeepFactors and our network architecture is shown in Fig.~\ref{fig:network}. The top network is a U-Net~\cite{ronneberger2015u} that takes as input a gray-scale image concatenated with a sparse depth and reprojection error maps from the sparse SLAM system. Similary to CodeSLAM and DeepFactors, the depth and reprojection error values are normalized to the range [0, 1] using the proximity parametrization ~\cite{bloesch2018codeslam}. The network uses the input to compute deep features by applying multiple convolution layers and outputs the depth uncertainty prediction. As Fig.~\ref{fig:rep_demo} shows, sparse SLAM landmarks can contain outliers and trusting all points equally can cause spikes in the predicted depth map. We observe that outlier points tend to have a higher average reprojection error value compared to inliers. Hence, we include the sparse reprojection error map in the input of the U-Net to provide the network with per-point uncertainty. The bottom network is a VAE conditioned by the decomposed features. The network generates latent code ${\bf{c}}$, dense depth prediction $D$  and uncertainty map ${b}$. The loss function of the network consists of a depth reconstruction loss and a KL divergence loss. The depth reconstruction loss is defined as:

\begin{eqnarray}
\sum_{{\bf{x}}\in{\Omega}} \frac{\|\: {D[{\bf{x}}]} - {D_{gt}}[{\bf{x}}] \:\|}{b[{\bf{x}}]} + log(b[{\bf{x}}]),
\end{eqnarray}

where ${\bf{x}}$ is a 2D pixel coordinate on the image plane ${\Omega}$ and ${D_{gt}}$ is the ground truth depth image.

\subsection{Training Procedure}

\begin{figure}[h]
    \centering
		\includegraphics[width=8.0cm]{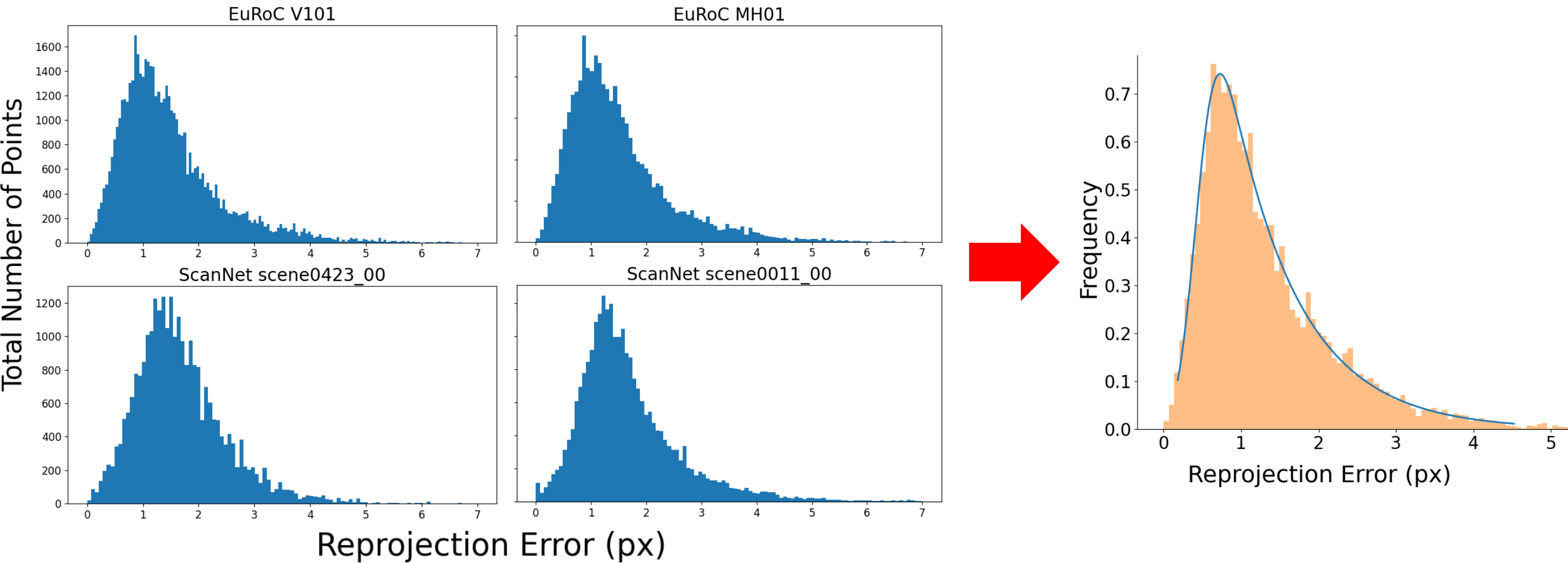}
		\caption{
            {~\bf{Left: }}The histogram of ORB-SLAM reprojection errors over different sequences. We can observe that the distribution is similar between sequences.{~\bf{Right: }}		    Probabilistic Model Fitting of the reprojection error histogram. We model the distribution by exponential Gaussian distribution (K=4.31, Mean=0.44, Sigma=0.20)
		}
		\label{fig:rep_histgram}
	
\end{figure}

We used $\sim0.4$M images from the ScanNet dataset \cite{dai2017scannet} following the official training/test split. ScanNet contains a variety of indoor RGBD images taken with a Kinect camera. To generate the sparse depth input, we run ORB keypoint detection on the color image and use these detections to select 1000 keypoints from each Kinect depth map. The input of the VAE is a raw Kinect depth image, and we use the depth map obtained by rendering the given 3D scene model from the current pose as ground truth depth for the loss function. Generating a reprojection error map is less trivial as it would not be feasible to run a SLAM system over the entire ScanNet dataset due to the long processing time. Instead, we propose a method for simulating reprojection errors during training. We observe that the distribution of reprojection errors from ORB-SLAM3 is similar between different sequences and closely resembles an exponential Gaussian distribution (Fig.~\ref{fig:rep_histgram}). We propose that adding noise to the sparse depth map so that resulting reprojection errors follow this distribution allows us to accurately simulate the properties of ORB-SLAM3.

\begin{figure}[h]
    \centering
		\includegraphics[width=8.0cm]{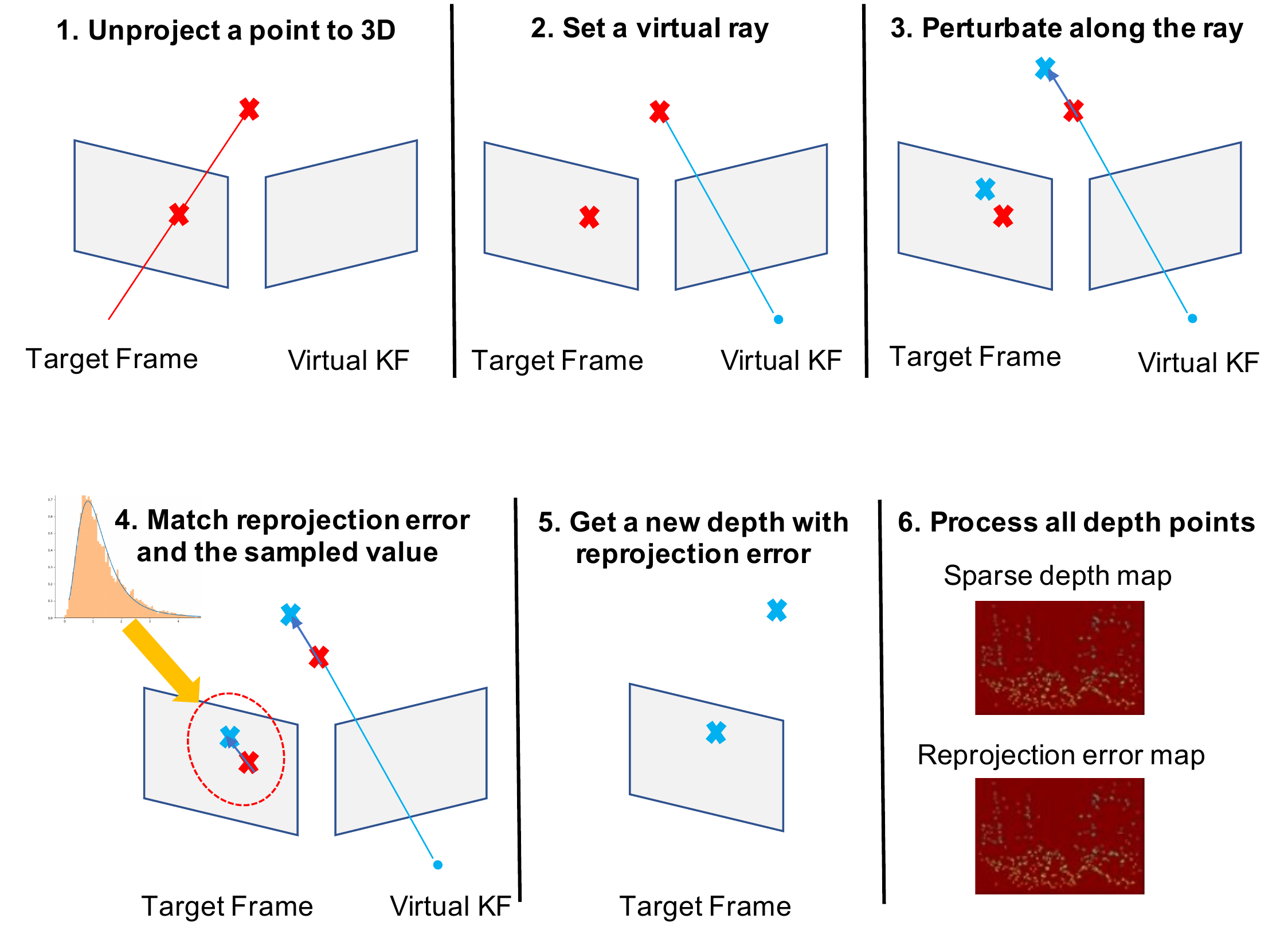}
		\caption{ Procedure to simulate per-point noise following the modeled reprojection error distribution described in in Fig.~\ref{fig:rep_histgram}.
		}
		\label{fig:gen_data}
	
\end{figure}

The dataset creation steps are detailed in Fig.~\ref{fig:gen_data}. Firstly, for every frame in the dataset (``reference frame''), we randomly select one neighboring frame which is within 2 meters from the reference (``virtual keyframe''). Then, we unproject points from the reference frame to 3D and compute the line that connects each 3D point to the virtual keyframes' camera center. For every point, we sample a reprojection error to simulate from the exponential Gaussian distribution. We use this to perturb a point's location along this line so that the distance between the 2D projection of the perturbed point and the 2D location of the original point on the reference frame matches this sampled reprojection error value.
We apply this process for the all sparse depth measurements and use the perturbed sparse depth map and reprojection error as input to the U-Net.  We trained the network for 10 epochs with a learning rate of 0.0001, image size of 256 × 192, code size 32 and using the Adam Optimizer.

\section{Multi-View Optimization}
\label{section:multiviewOpt}
\begin{figure}[h]
    \centering
        \fbox{
		\includegraphics[width=8.0cm]{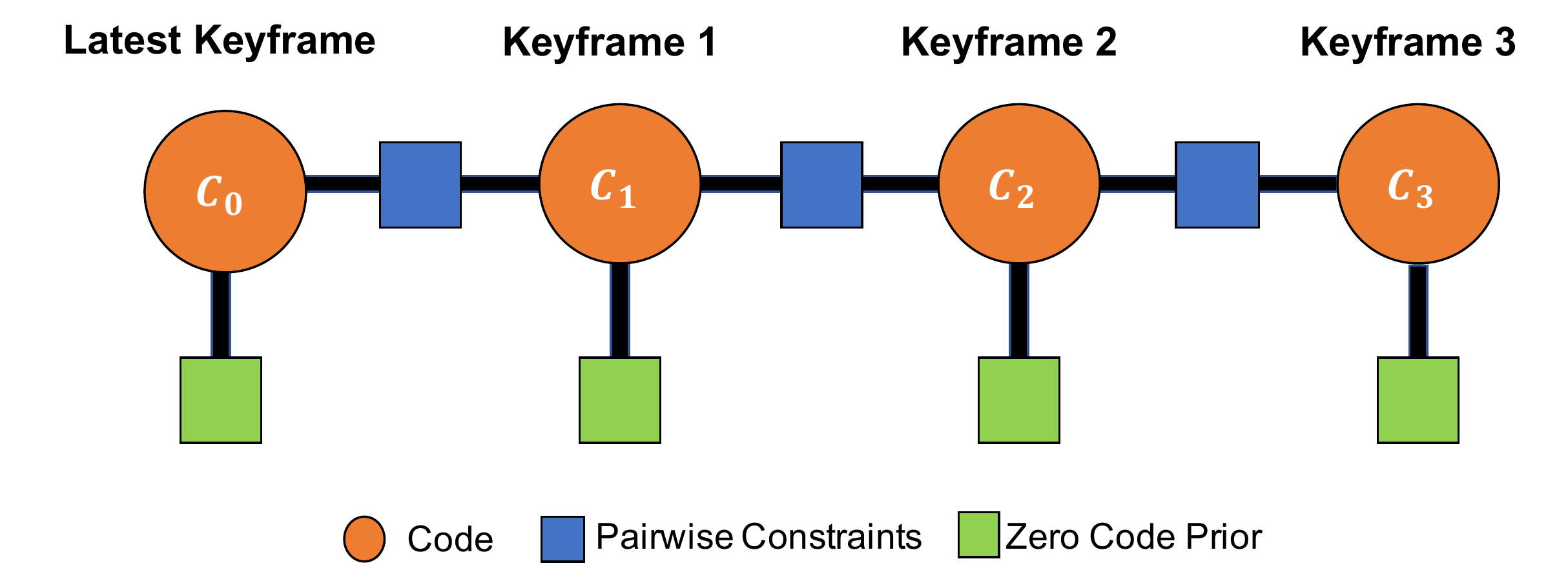}
		}
		\caption{ 
		Factor graph formulation of multi-view optimization. The optimization window consists of 4 keyframes and constraints are added between consecutive pairs of frames. We include a zero-code prior for regularization.}
		
		\label{fig:factorgraph}
	
\end{figure}

We improve the accuracy and consistency of per-frame depth predictions by running multi-view dense bundle adjustment (Fig.~\ref{fig:multiview}). Following the methodology proposed in DeepFactors, we optimize depth images on a compact manifold generated by the conditional VAE. Using the code representation, we minimize a set of different objective functions between consecutive frames that ensure consistency of observations from the overlapping frames and find the best estimate of scene geometry. We formulate this problem as a factor graph (Fig.~\ref{fig:factorgraph}), where the variables are the per-image codes and the factors are the error terms described next . We use  DeepFactors' optimizer implemented using the GTSAM library~\cite{dellaert2012factor} to solve this problem. The following sections describe these factors in greater detail. We use Huber norm on all factors as a robust cost function.

\subsection{Photometric Factor}
\label{subsec:photometricFactor}
The Photometric Factor optimises for photometric consistency between images $I_i$ and $I_j$. It computes the sum of pixel-wise squared differences formulated as:
\begin{eqnarray}
_{photo}{E^{ij}} ({\bf{c}}_i) = \sum_{{\bf{x}}\in{{\Omega}}} \|\:I_i [{\bf{x}}] - I_j[{\omega}_{ji}({\bf{x}},{\bf{c}}_i,I_i)]\:\|^2 ~,
\end{eqnarray}
where ${\bf{x}}$ is a 2D pixel coordinate on the image plane ${\Omega}$ and ${\omega_{ji}}$ is a function which warps this coordinate from frame $i$ to frame $j$:
\begin{eqnarray}
\omega_{ji}({\bf{x}},{\bf{c}}_i,I_i) = \pi({\bf{T}}_{ji} (\pi^{-1}({\bf{x}}, D_i[{\bf{x}}]))) ~.
\end{eqnarray}
The $\pi$ function projects a 3D point onto a 2D plane while $\pi^{-1}$ is the inverse operation. $D_i = D({\bf{c}}_i , I_i, S_i, R_i)$ is the depth map decoded from code ${{\bf{c}}_i} \in \mathbb{R}^{32}$, intensity image $I_i$, sparse depth image $S_i$ and sparse reprojection error image $R_i$. ${{\bf{T}}_{ji}} \in {\mathbb{SE}(3)}$ is the 6 DoF transformation from frame $i$ to frame $j$.

\subsection{Reprojection Factor}
The Reprojection Factor minimizes the reprojection error of matched keypoints between frames. Similar to DeepFactors, we use BRISK as a keypoint detector and descriptor. The objective function is defined as:
\begin{eqnarray}
_{rep}{E^{ij}}({\bf{c}}_i) = \sum_{{\bf{x,y}}\in{{M}_{ij}}} \|\:{\omega}_{ji}({\bf{x}},{\bf{c}}_i,I_i) - {\bf{y}}\:\|^2
\end{eqnarray}
where $M_{ij}$ is a set of matched feature points between frames $i$ and $j$. {\bf{x}} and {\bf{y}} are their corresponding 2D image coordinates on frames $i$ and $j$ respectively.

\subsection{Sparse Geometric Factor}
The Sparse Geometric Factor directly measures the per-pixel difference between the depth maps after warping them to a common reference frame. Due to computational limitations, we uniformly sample a set of pixels on the image plane for which we compute this error ($\Omega'$). The factor helps to improve the consistency of textureless regions where the photometric and reprojection factors do not provide sufficient signal for optimization. It is formulated as:
\begin{eqnarray}
_{dpt}E^{ij} ({\bf{c}}_i) = \sum_{{\bf{x}}\in{{\Omega'}}}\|\:| {\bf{T}}_{ji}\pi^{-1}({\bf{x}},D_i[{\bf{x}}])|_z - D_j[\hat{{\bf{x}}}]\:\|^2
~,
\end{eqnarray}

where $D_i$ and $D_j$ are the depth maps decoded for frames $i$ and $j$ as described in Section~\ref{subsec:photometricFactor}, $\hat{{\bf{x}}} = {\omega}_{ji}({\bf{x}}, {\bf{c}}_i, I_i)$ and $|{\bf{v}}|_z$ denotes taking the $z$ component of the vector ${\bf{v}}$.

\section{Experimental Results}
We trained our model with a ScanNet training split using the Python TensorFlow library. We export the trained model to the C++ SLAM system using the TensorFlow C API. Similarly to DeepFactors, our dense optimization module has been implemented in C++ with CUDA. We run all experiments in an Intel Core i9-10900 CPU at 2.8 GHz and NVIDIA GTX 3080 GPU. We have used an image resolution of 256 × 192 for dense mapping. We evaluate our system on the ScanNet and EuRoC MAV datasets.

\subsection{Quantitative Depth Accuracy Evaluation}
We compared our method against DeepFactors \cite{czarnowski2020deepfactors} (where depth image predictions are only conditioned on intensity images) and the depth completion method proposed by Ma ~\emph{et al.} \cite{ma2018sparse} (single-view depth completion given a set of sparse points). We trained the network model from Ma ~\emph{et al.} \cite{ma2018sparse} on the ScanNet training set with 1000 depth points per image.
As an evaluation metric, we measured mean absolute error (MAE) and root mean square error (RMSE) between the predicted and the ground truth depth images. We also evaluate the depth prediction of our method under different configurations, including with and without reprojection error conditioning or multi-view optimisation.

\subsubsection{ScanNet Dataset}
\begin{figure}[h]
    \centering
		\includegraphics[width=8.0cm]{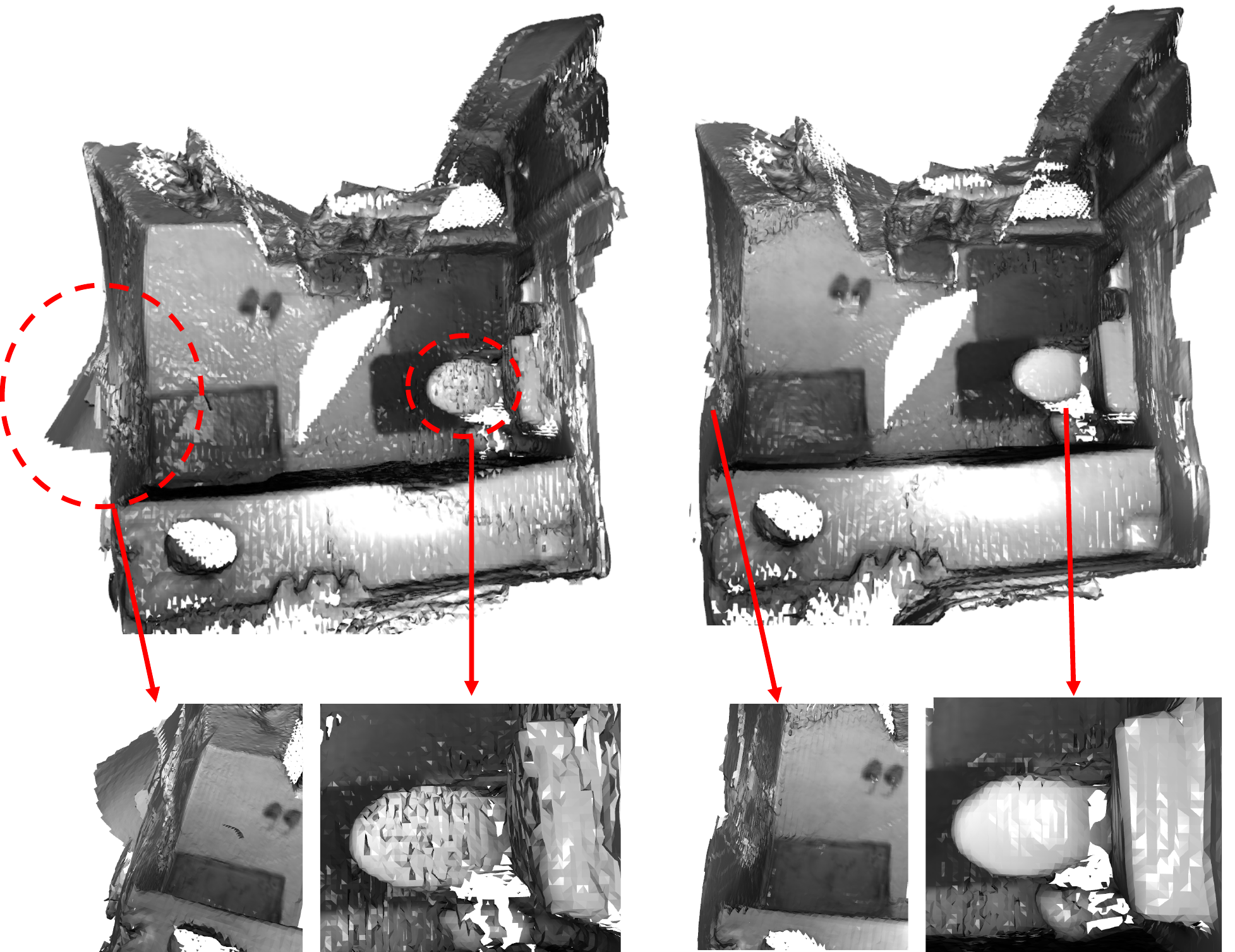}
		\caption{
    		TSDF fusion of predicted depth maps (\textbf{left:} without multi-view optimization, \textbf{right:} with multi-view optimization). As highlighted in the red circles, multi-view optimization helps to increase depth map accuracy and consistency.
		}
		\label{fig:tsdf_multiview}
	
\end{figure}

We run ORB-SLAM3 in RGBD mode to estimate per-keyframe camera posese, sparse depth and reprojection error maps consisting of 200 to 1000 points per frame. We run our system on the test sequences of this dataset and evaluate the depth prediction accuracy against ground truth depth images from the renderings of the provided 3D models.  We report our results in Table~\ref{table:scannet}. Our method outperforms the other dense prediction networks in all sequences. Multi-view optimization further improves our single-view dense prediction by approximately $10\%$. This can be seen in Fig.~\ref{fig:tsdf_multiview}, where we show a comparison of TSDF fusion of depth images estimated with and without multi-view optimization. Optimization leads to more accurate and also more consistent depth maps that produce higher quality 3D reconstructions. On ScanNet, conditioning on reprojection error does not lead to significant performance improvement due to ORB-SLAM3 generally working well and not producing gross outliers.

\setlength{\tabcolsep}{0.55mm}
\begin{table}[h]
\caption{Depth Prediction results of the ScanNet dataset (in meters)} 
\label{table:scannet}
\centering
\fbox{
    \scalebox{0.75}[0.85]{
    \begin{tabular}{p{2.5em}|c|ccccccccc}
    \hline
      \multicolumn{3}{c}{} & 0100\_00 & 0406\_02 & 0382\_01 & 0084\_00 & 0702\_02 &0406\_01 &0695\_03\\
    \hline\hline
    \multicolumn{2}{c|}{\begin{tabular}{c} ORB Points \\(sparse)\end{tabular}}
    &\begin{tabular}{c} MAE\\ RMSE\end{tabular} 
    &\begin{tabular}{c}.028\\.065\end{tabular} &\begin{tabular}{c}.031\\.092\end{tabular} &\begin{tabular}{c}.068\\.134\end{tabular} &\begin{tabular}{c}.063\\.124\end{tabular} &\begin{tabular}{c}.032\\.087\end{tabular}
    &\begin{tabular}{c}.063\\.133\end{tabular}
    &\begin{tabular}{c}.072\\.120\end{tabular}
    \\
    \hline
    \hline
    \multicolumn{2}{c|}{DeepFactors~\cite{czarnowski2020deepfactors}} 
    &\begin{tabular}{c} MAE\\ RMSE\end{tabular} 
    &\begin{tabular}{c} .185 \\ .220\end{tabular} &\begin{tabular}{c}.177\\.203\end{tabular} &\begin{tabular}{c}.358\\.432\end{tabular} &\begin{tabular}{c}.237\\.276\end{tabular} &\begin{tabular}{c}.368\\.397\end{tabular}
    &\begin{tabular}{c}.267\\.398\end{tabular}
    &\begin{tabular}{c}.291\\.408\end{tabular}
    \\
    
    \hline
    \multicolumn{2}{c|}{Ma~\emph{et al.}~\cite{ma2018sparse}}
    & \begin{tabular}{c} MAE\\ RMSE\end{tabular}
    &\begin{tabular}{c}.141\\.285\end{tabular} &\begin{tabular}{c}.088\\.208\end{tabular} &\begin{tabular}{c}.140\\.284\end{tabular} &\begin{tabular}{c}.199\\.357\end{tabular} &\begin{tabular}{c}.123\\.235\end{tabular}
    &\begin{tabular}{c}.251\\.321\end{tabular}
    &\begin{tabular}{c}.272\\.337\end{tabular}
    \\
    \hline
    
    \multirow{7}{*}{Ours}&\begin{tabular}{c} w/o rep error \\ \& multiview opt. \end{tabular}
    &\begin{tabular}{c} MAE\\ RMSE\end{tabular}
    &\begin{tabular}{c} .050 \\ .102\end{tabular} &\begin{tabular}{c}.066\\.140\end{tabular}  &\begin{tabular}{c}.123\\.277\end{tabular}  &\begin{tabular}{c}.069\\.170\end{tabular}  &\begin{tabular}{c}.069\\.122\end{tabular}
    &\begin{tabular}{c}.070\\.141\end{tabular}  &\begin{tabular}{c}.086\\.127\end{tabular}
    \\
    \cline{2-10}
    &\begin{tabular}{c}w/o rep error\end{tabular}
    &\begin{tabular}{c} MAE\\ RMSE\end{tabular}
    &\begin{tabular}{c}\bf.044\\\bf.098\end{tabular} 
    &\begin{tabular}{c}.061\\.141\end{tabular}
    &\begin{tabular}{c}.116\\.273\end{tabular}
    &\begin{tabular}{c}\bf{.067}\\\bf{.165}\end{tabular} 
    &\begin{tabular}{c}.061\\.120\end{tabular}
    &\begin{tabular}{c}.064\\.139\end{tabular} 
    &\begin{tabular}{c}\bf{.079}\\.127\end{tabular}
    \\
    \cline{2-10}
    
    &\begin{tabular}{c}w/o \\ multiview opt.\end{tabular}
    &\begin{tabular}{c} MAE\\ RMSE\end{tabular}
    &\begin{tabular}{c}.059\\.110\end{tabular} 
    &\begin{tabular}{c}.068\\.142\end{tabular}
    &\begin{tabular}{c}.129\\.266\end{tabular}
    &\begin{tabular}{c}.089\\.181\end{tabular}
    &\begin{tabular}{c}.073\\.123\end{tabular} 
    &\begin{tabular}{c}.069\\.141\end{tabular}
    &\begin{tabular}{c}.097\\.143\end{tabular} 
    \\
    \cline{2-10}

    &\begin{tabular}{c}Full\end{tabular}
    &\begin{tabular}{c} MAE\\ RMSE\end{tabular}
    &\begin{tabular}{c}.046\\.100\end{tabular}
    &\begin{tabular}{c}\bf.060\\\bf.136\end{tabular}
    &\begin{tabular}{c}\bf{.115}\\\bf.260\end{tabular}
    &\begin{tabular}{c}.073\\.172\end{tabular}
    &\begin{tabular}{c}\bf.059\\\bf.113\end{tabular}
    &\begin{tabular}{c}\bf.063\\\bf.137\end{tabular}
    &\begin{tabular}{c}\bf.079\\\bf.125\end{tabular}
    \\
    \hline
    \end{tabular}
    }
}
\end{table}

\begin{figure}
\centering
\fbox{
	\includegraphics[width=8.0cm]{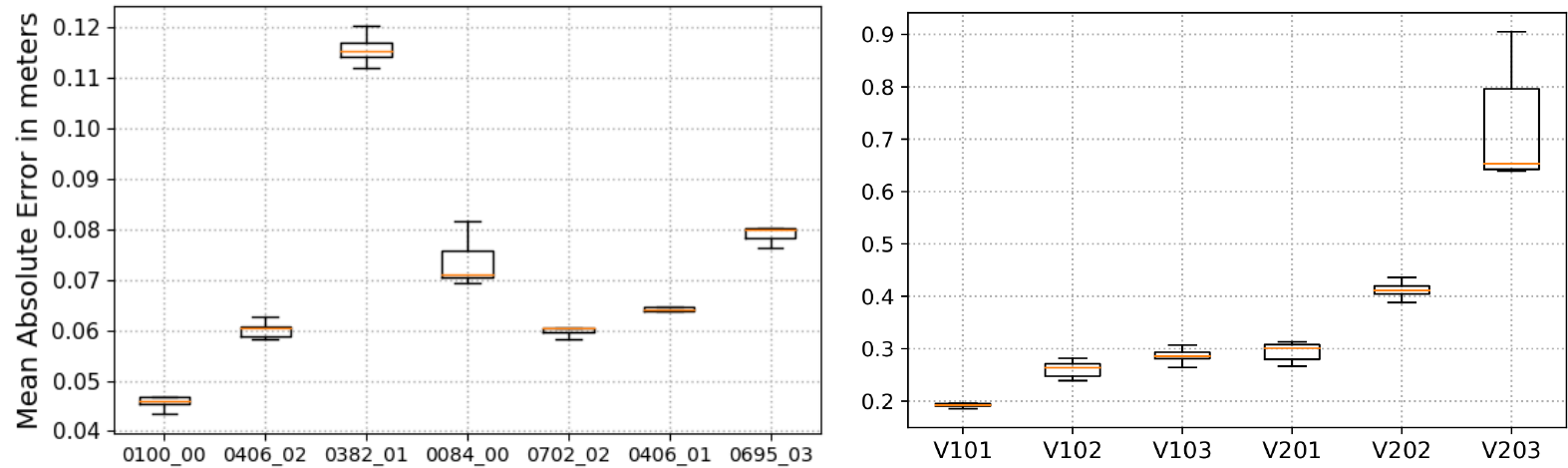}
	}
		\caption{
		Boxplot of our method (Full) on 2 datasets (\textbf{left:} ScanNet, \textbf{right:} EuRoC). The vertical axis represents Mean Absolute Error of depth prediction. We run our method 5 times on each sequence.
		\label{fig:boxplot}
		}
\end{figure}

\subsubsection{EuRoC MAV Dataset}
\label{sec:euroc}

\begin{figure}[h]
    \centering
		\includegraphics[width=8.5cm]{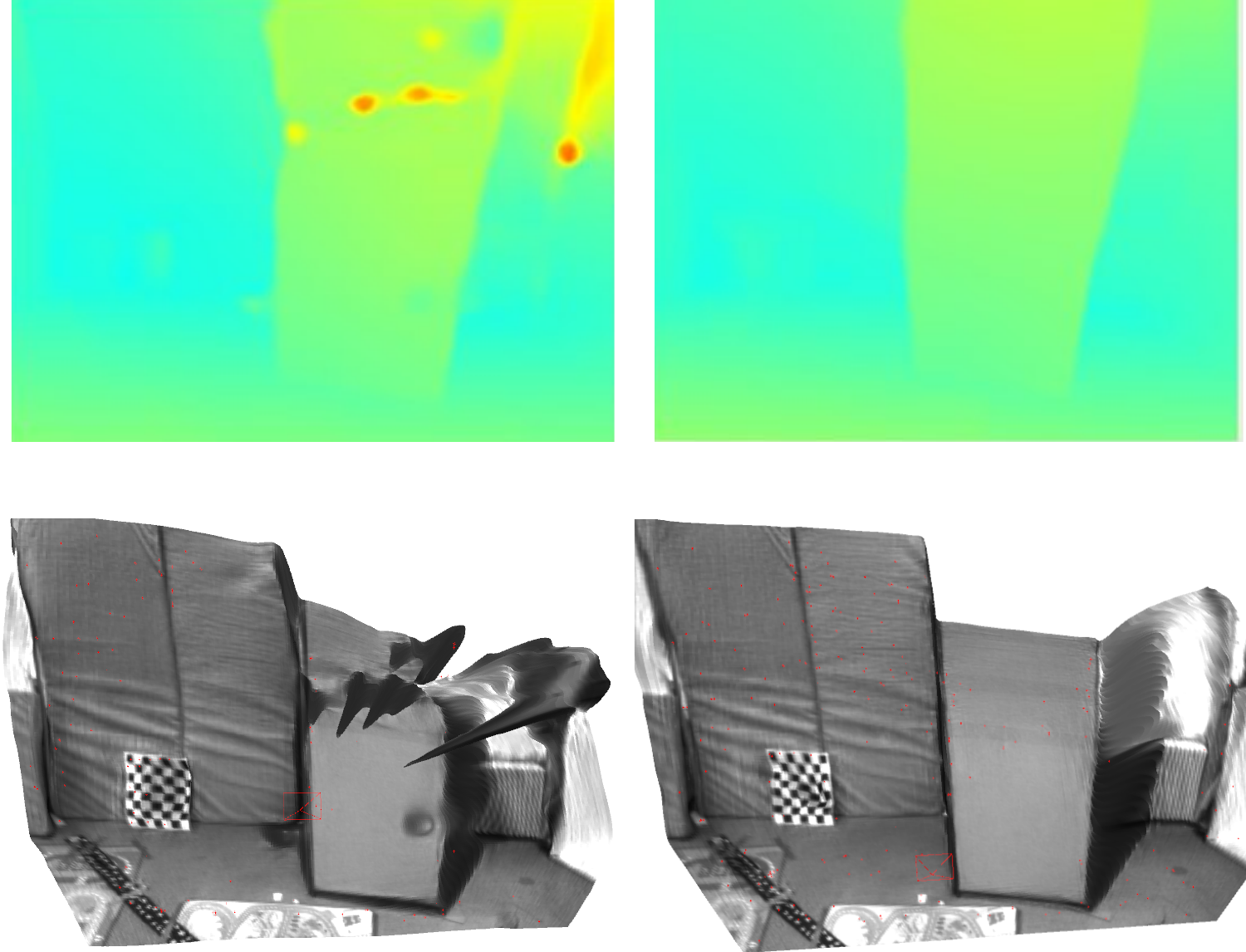}
		\caption{ 
		A 2D/3D visualization of a depth prediction sample on the EuRoC dataset (\textbf{left:} without reperojection error conditioning, \textbf{right:} with reperojection error conditioning). Reprojection error conditioning helps in filtering out gross outliers.
		}
		\label{fig:rep_change}
\end{figure}

We also evaluate our method on the EuRoC MAV dataset \cite{burri2016euroc} by running ORB-SLAM3 in visual-inertial mode. We generate ground truth depth maps by
rendering the LiDAR data onto each frame. While the sequences from the ScanNet training set contain small rooms with furniture, the EuRoC dataset has much larger scenes with very different objects. We report reconstruction error results for different configurations of our system in Table~\ref{table:euroc}.
These error values are higher compared to those obtained on the ScanNet dataset. We think this happens because the network learns the sizes and shapes of objects typically found in the apartments of the ScanNet dataset, while the environment in the EuRoC dataset is a large laboratory with technical equipment. However, compared to DeepFactors (which uses only intensity image conditioning), our method is less affected by the dataset domain change. This demonstrates that conditioning on sparse points improves the generalization capability of the network. Regarding reprojection error conditioning, Fig.~\ref{fig:rep_change} shows an example of its importance for filtering out gross outliers. Outliers are more common in EuRoC as the dataset contains more texture-less walls and the 3D location of features is more difficult to calculate accurately.
Fig.~\ref{fig:boxplot} shows errors over multiple trials on both ScanNet and EuRoC sequences to indicate the statistical significance of our results.

\setlength{\tabcolsep}{1.1mm}
\begin{table}[h]
\caption{Depth Prediction Results of the EuRoC dataset (in meters)}
\label{table:euroc}
\centering
\fbox{
    \scalebox{0.80}[0.85]{
    \begin{tabular}{p{2.5em}|c|ccccccc}
    \hline
    \multicolumn{3}{c}{}& V101 & V102 & V103  & V201  & V202  & V203\\
    \hline\hline
    \multicolumn{2}{c|}{\begin{tabular}{c} ORB Points \\(sparse)\end{tabular}}&
    \begin{tabular}{c} MAE\\ RMSE\end{tabular} &
    \begin{tabular}{c}.115\\.283\end{tabular} &\begin{tabular}{c}.177\\.381\end{tabular} & \begin{tabular}{c}.216\\.405\end{tabular} & \begin{tabular}{c}.183\\.374\end{tabular} & \begin{tabular}{c}.291\\.617\end{tabular} &
    \begin{tabular}{c}.610\\.801\end{tabular}
    \\
    
    \hline\hline
    \multicolumn{2}{c|}{\begin{tabular}{c}DeepFactors~\cite{czarnowski2020deepfactors}\end{tabular}}
    &\begin{tabular}{c} MAE\\ RMSE\end{tabular}
    &\begin{tabular}{c}.842\\1.05\end{tabular} &\begin{tabular}{c}.875\\1.03\end{tabular}
    &\begin{tabular}{c}.833\\.940\end{tabular}
    &\begin{tabular}{c}.859\\1.02\end{tabular}
    &\begin{tabular}{c}1.29\\1.53\end{tabular}
    &\begin{tabular}{c}1.08\\1.25\end{tabular}
    \\
    \hline
    \multicolumn{2}{c|}{\begin{tabular}{c}Ma~\emph{et al.}~\cite{ma2018sparse}\end{tabular}}
    &\begin{tabular}{c} MAE\\ RMSE\end{tabular}
    &\begin{tabular}{c}.495\\.598\end{tabular} &\begin{tabular}{c}.509\\.611\end{tabular}
    &\begin{tabular}{c}.492\\.595\end{tabular}
    &\begin{tabular}{c}.486\\.610\end{tabular}
    &\begin{tabular}{c}.504\\.660\end{tabular}
    &\begin{tabular}{c}.871\\1.08\end{tabular}
    \\
    \hline
    
    \multirow{7}{*}{Ours}&\begin{tabular}{c} w/o rep error \\ \& multiview opt. \end{tabular}
    &\begin{tabular}{c} MAE\\ RMSE\end{tabular}
    &\begin{tabular}{c}.280\\.435\end{tabular} 
    &\begin{tabular}{c}.334\\.446\end{tabular}
    &\begin{tabular}{c}.354\\.460\end{tabular}
    &\begin{tabular}{c}.390\\.572\end{tabular} 
    &\begin{tabular}{c}.533\\.782\end{tabular}
    &\begin{tabular}{c}.790\\1.14\end{tabular}
    \\
    \cline{2-9}
    &\begin{tabular}{c}w/o rep error\end{tabular}
    &\begin{tabular}{c} MAE\\ RMSE\end{tabular}
    &\begin{tabular}{c}.231\\.417\end{tabular} 
    &\begin{tabular}{c}.289\\.419\end{tabular}
    &\begin{tabular}{c}.356\\.457\end{tabular}
    &\begin{tabular}{c}.336\\.551\end{tabular} 
    &\begin{tabular}{c}.515\\.764\end{tabular}
    &\begin{tabular}{c}.780\\1.10\end{tabular}
    \\
    \cline{2-9}
    
    &\begin{tabular}{c}w/o \\ multiview opt.\end{tabular}
    &\begin{tabular}{c} MAE\\ RMSE\end{tabular}
    &\begin{tabular}{c}.296\\.450\end{tabular} 
    &\begin{tabular}{c}.358\\.479\end{tabular}
    &\begin{tabular}{c}.412\\.528\end{tabular}
    &\begin{tabular}{c}.373\\.566\end{tabular} 
    &\begin{tabular}{c}.505\\.748\end{tabular}
    &\begin{tabular}{c}.725\\1.00\end{tabular}
    \\
    \cline{2-9}

    &\begin{tabular}{c}Full\end{tabular}&
    \begin{tabular}{c} MAE\\ RMSE\end{tabular}&
    \begin{tabular}{c}\bf{.192}\\\bf.381\end{tabular}&
    \begin{tabular}{c}\bf.259\\\bf.369\end{tabular} &
    \begin{tabular}{c}\bf{.283}\\\bf.407\end{tabular}&
    \begin{tabular}{c}\bf{.290}\\\bf.428\end{tabular}&
    \begin{tabular}{c}\bf.415\\\bf.655\end{tabular} &
    \begin{tabular}{c}\bf{.686}\\\bf.952\end{tabular}
    \\
    \hline
    \end{tabular}
    }
}
\end{table}

\subsubsection{Runtime measurement}
Table ~\ref{table:runtime} shows the measured average time over the EuRoC and ScanNet dataset in milliseconds. Dense depth prediction times are reported per frame while multi-view optimization times are reported for formulations with 4 keyframes. We used the TensorFlow C++ API to run the trained model, but as Table~\ref{table:runtime} shows we found that running the densification network via the Python API is 20 times faster due to a better model loading procedure. We believe this speedup could also be replicated for our C++ system with an equivalent model loading process. Our approach can predict 4 dense depth images and refine them through multi-view optimization at a rate of around 1Hz. This frequency is sufficient for map updates as long as the camera motion is not aggressive. 

\begin{table}[h]
\caption{Mean timing results (milliseconds).}
\label{table:runtime}
\centering
\fbox{
\scalebox{1.0}[1.0]{
\begin{tabular}{|l|l|}
\hline
\begin{tabular}{l}Dense Prediction\end{tabular} &\begin{tabular}{l}235 \scriptsize{(C++ API)} \\ 11 \scriptsize{(Python API)}\end{tabular}\\
\hline
\begin{tabular}{l}Multi-view Optimization\end{tabular}&
\begin{tabular}{l}170\end{tabular}\\
\hline

\end{tabular}
}
}

\end{table}
\subsubsection{Limitations}
As we discussed in Sec. \ref{sec:euroc}, the performance of the network can be affected by large domain changes between training and testing data. This is mitigated to some extent by sparse points conditioning which acts as an online scale adjustment.
Also, since we assume high quality points and camera poses given by the SLAM system, our method cannot handle scenes where SLAM fails due to low texture or very fast camera motion. However, as long as visual tracking can work, the SLAM system normally provides around 200 to 1000 points per-frame which is sufficient for depth completion to work well.

\subsection{Qualitative Map Evaluation}
\subsubsection{Local Mapping}
Our dense mapper continuously refines a keyframe window which consists of the latest keyframe and its top 3 covisibility keyframes by minimizing the error functions described in Section~\ref{section:multiviewOpt}. This process improves consistency for depth maps in overlapping frames.

\begin{figure}[h]
    \centering
    \fbox{
		\includegraphics[width=7.5cm]{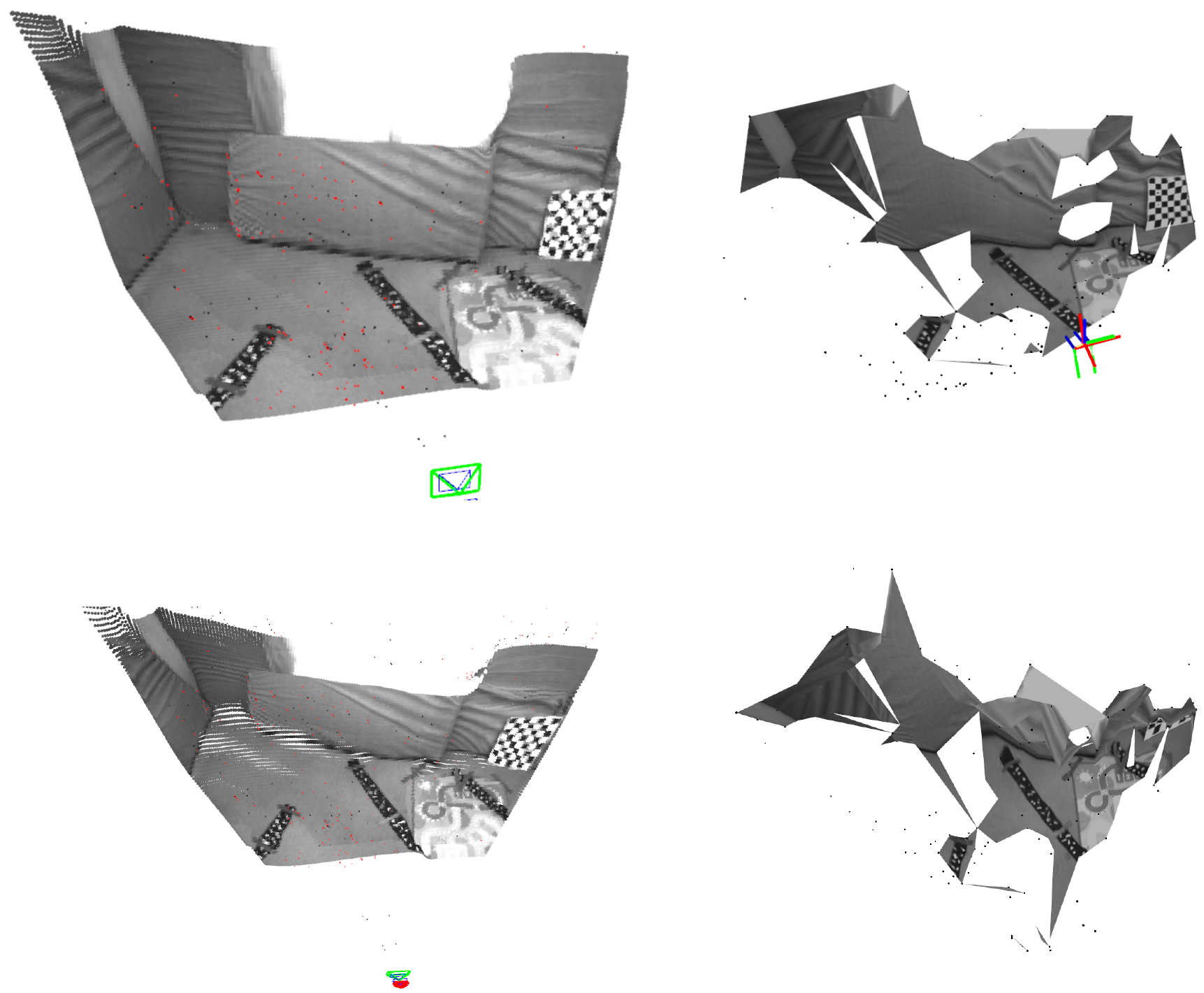}
	}
		\caption{
		    An example of dense local mapping  (\textbf{left}: ours, \textbf{right}: Delaunay meshing by Kimera~\cite{rosinol2020kimera}. While both methods rely on visual-inertial SLAM, our method generates more complete and smooth local maps.}
		\label{fig:local_mapping}
	
\end{figure}

We compare this procedure to Kimera~\cite{rosinol2020kimera}, a real-time geometric system which generates meshes using 2D Delaunay triangulation and sparse SLAM points. Fig.~\ref{fig:local_mapping} shows a comparison between maps generated by our system and those from Kimera. Using a network to learn about shapes and sizes of objects in indoor spaces leads to more accurate and smooth results compared to purely geometric approaches.

\subsubsection{Globally Consistent Depth Map Fusion}
\begin{figure}[h]
        \centering
        \fbox{
		    \includegraphics[width=7.5cm]{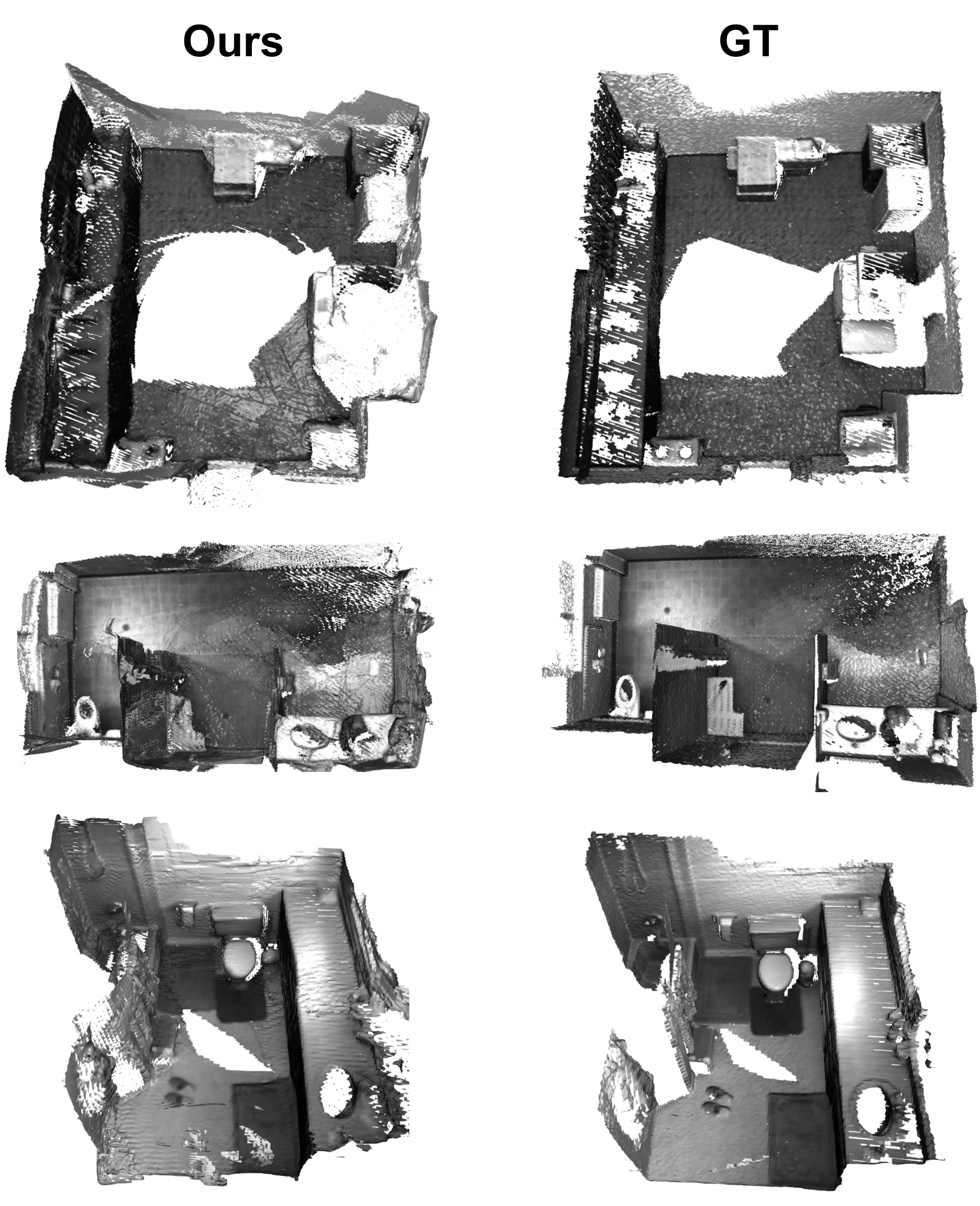}
		}
				\caption{
    				Example 3D reconstructions on the ScanNet test set from our system (left) compared to ground truth (right). We generate globally consistent 3D models using TSDF fusion of depth maps from our system and camera poses from the SLAM  system.}
		\label{fig:tsdffusion}
        
\end{figure}

\begin{figure}[h]
    \centering
		\includegraphics[width=8.0cm]{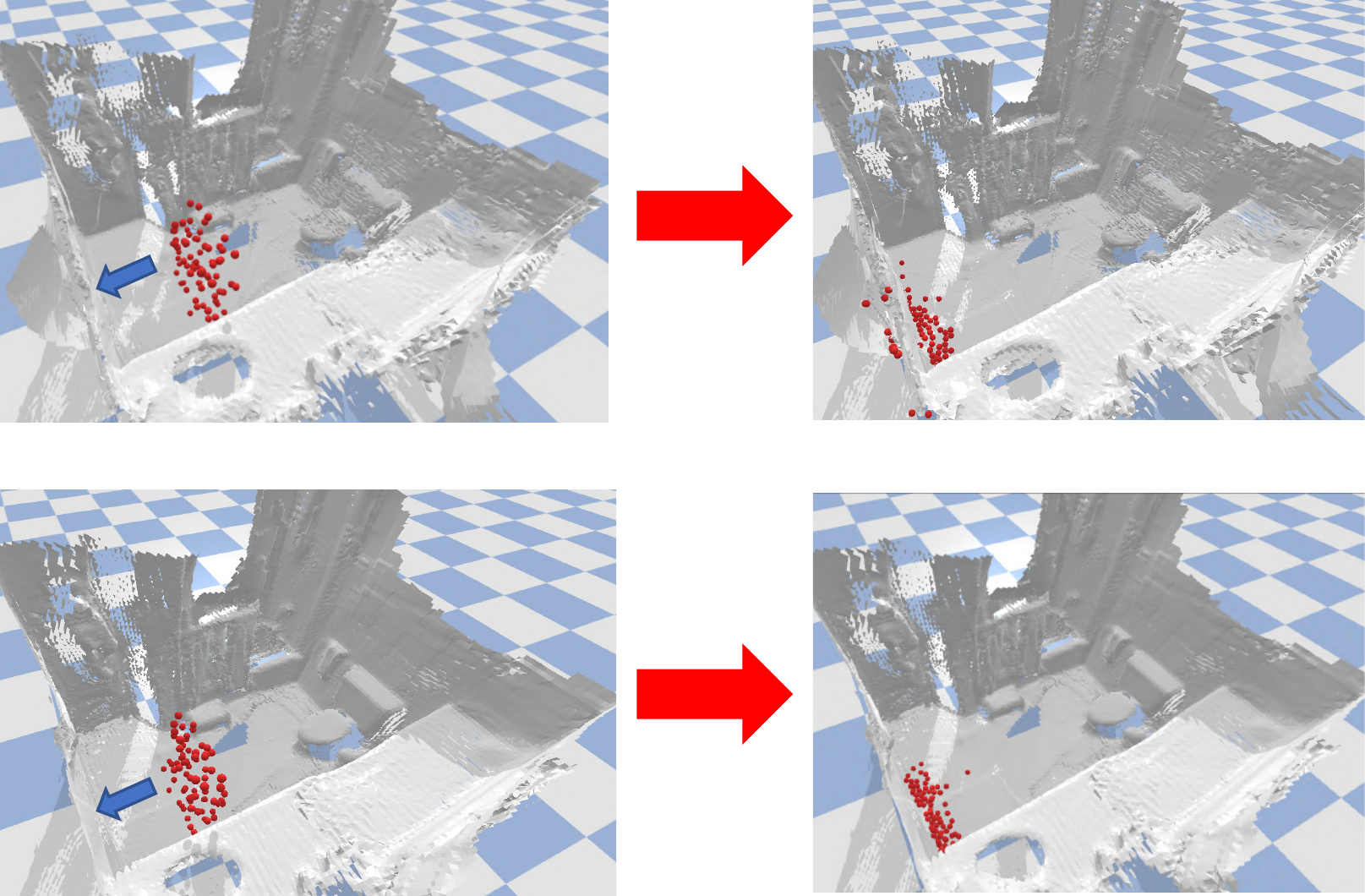}
		\caption{
		    A physics simulation example using the dense 3D model from TSDF fusion -- generated without ({\bf{top}}) and with ({\bf{bottom}}) multi-view optimization, where red balls are thrown towards the wall. In the top 3D model the wall is not straight and the balls sink into it, but in the bottom model the balls bounce off the wall correctly thanks to accurate geometry enabled by multi-view optimization.
		}
		\label{fig:physics}
	
\end{figure}

Since the SLAM thread (ORB-SLAM3) provides accurate and globally consistent camera poses, we can not only refine our dense prediction but also fuse depth maps into a globally consistent dense 3D model. Fig.~\ref{fig:tsdffusion} shows examples of offline volumetric TSDF fusion of depth maps from CodeMapping on scenes from the ScanNet test set. This is done after the SLAM system has processed the entire sequence. These 3D models are essential for tasks which require rich geometry information such as obstacle avoidance, AR/VR and physics simulation. In Fig.~\ref{fig:physics} we carry out a physics simulation experiment by taking TSDF models generated with and without multi-view optimization and we simulate throwing a set of balls towards a wall. The example illustrates the practical importance of methods which generate not only dense but consistent depth maps.

\subsubsection{Pure Monocular Dense Mapping}
\begin{figure}[h]
        \centering
        \fbox{
		    \includegraphics[width=8.0cm]{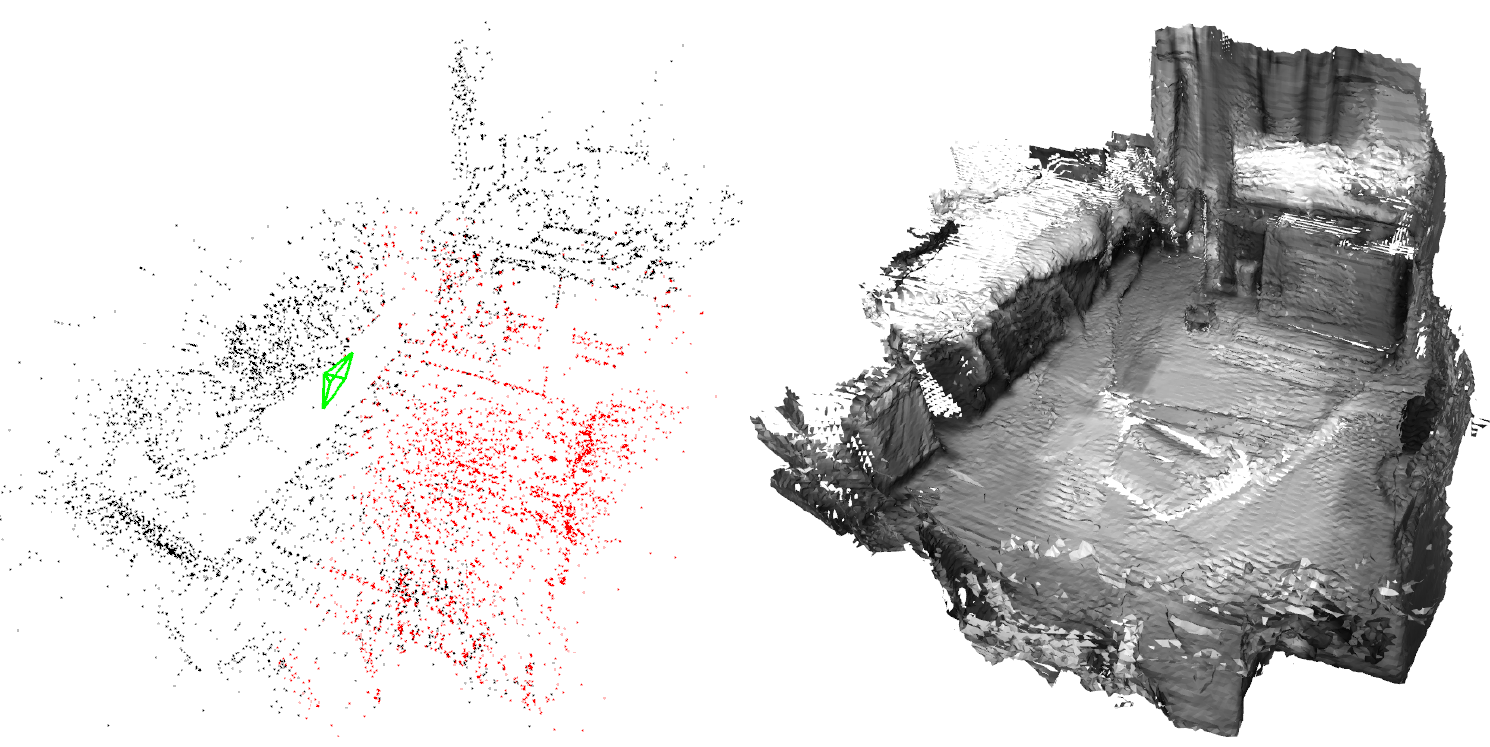}
		}
				\caption{
    				Example 3D reconstruction from pure monocular ORB-SLAM (\textbf{left}: monocular SLAM result, \textbf{right}: our dense mapping result). The scene represents an apartment sequence which we manually collected.
			}
		\label{fig:monomapping}
\end{figure}

Although the results presented so far were obtained using metric SLAM, our method could also be applied to a scale-less monocular SLAM result. We define the scale factor as the median of the depth maps of the training data divided by the median of the depth maps from monocular ORB-SLAM. We multiply the depth maps and translational components of the camera poses provided by monocular SLAM by this scale factor and obtain 3D reconstructions of promising quality (Fig.~\ref{fig:monomapping}).

\section{Conclusions}
In this paper, we introduced a novel real-time multi-view dense mapping method which complements sparse SLAM systems. We first proposed a novel VAE architecture conditioned on image intensity, sparse depth and reprojection error maps, which performs uncertainty-aware depth completion. We also proposed an approach to integrate the VAE into an existing keyframe-based sparse SLAM system using multi-threading. This flexible integration allows us to run dense bundle adjustment without delaying the main SLAM process. We evaluate our system on two public datasets and demonstrate that our method performs better than other dense depth prediction methods, while multi-view optimization further improves accuracy and consistency. Our mapper provides not just locally but also globally consistent depth maps through fusion in a 3D global TSDF model. In future work, our mapper can be extended to use semantic segmentation and instance segmentation for more informative 3D mapping. 

{\small
\bibliographystyle{IEEEtran}
\bibliography{abbrev_short,egbib}
}

\end{document}